\documentclass[sigconf]{acmart}

\usepackage{url}            
\usepackage{booktabs}       
\usepackage{amsfonts}       
\usepackage{nicefrac}       
\usepackage{microtype}      
\usepackage{mathrsfs}
\usepackage{graphicx} 
\graphicspath{{plots/}} 
\DeclareMathOperator{\E}{\mathbb{E}}
\DeclareMathOperator*{\argmax}{argmax}

\usepackage{algorithm}
\usepackage{algpseudocode}
\usepackage{booktabs}       
\usepackage{pbox}           
\usepackage{geometry}
\usepackage[utf8]{inputenc} 
\usepackage[T1]{fontenc}    
\usepackage{hyperref}       
\usepackage{url}            
\usepackage{booktabs}       
\usepackage{amsfonts}       
\usepackage{nicefrac}       
\usepackage{microtype}      
\usepackage{bm}             
\usepackage{graphicx}       
\usepackage{array}          
\usepackage{booktabs}       
\usepackage{pbox}           
\usepackage{subcaption}     

\let\oldemptyset\emptyset

\graphicspath{{plots/}} 

\AtBeginDocument{%
  \providecommand\BibTeX{{%
    \normalfont B\kern-0.5em{\scshape i\kern-0.25em b}\kern-0.8em\TeX}}}

\setcopyright{none}
\copyrightyear{2021}
\acmYear{2021}

\acmConference[Marble-KDD 21]{Marble-KDD 21}{August 16, 2021}{Singapore}

\begin{document}

\title{Markov Decision Process modeled with Bandits for Sequential Decision Making in Linear-flow}


\author{Wenjun Zeng}
\email{zengwenj@amazon.com}
\affiliation{%
  \institution{Amazon.com}
 \city{Seattle}
 \state{Washington}
 \country{United States}
}
\author{Yi Liu}
\email{yiam@amazon.com}
\affiliation{%
  \institution{Amazon.com}
   \city{Seattle}
 \state{Washington}
 \country{United States}
}

\renewcommand{\shortauthors}{}

\begin{abstract}
For marketing, we sometimes need to recommend content for multiple pages in sequence. Different from general sequential decision making process, the use cases have a simpler flow where customers per seeing recommended content on each page can only return feedback as moving forward in the process or dropping from it until a termination state. We refer to this type of problems as sequential decision making in linear--flow. We propose to formulate the problem as an \textit{MDP with Bandits} where Bandits are employed to model the transition probability matrix. At recommendation time, we use Thompson sampling (TS) to sample the transition probabilities and allocate the best series of actions with analytical solution through exact dynamic programming. The way that we formulate the problem allows us to leverage TS's efficiency in balancing exploration and exploitation and Bandit's convenience in modeling actions' incompatibility. In the simulation study, we observe the proposed \textit{MDP with Bandits} algorithm outperforms Q-learning with $\epsilon$-greedy and decreasing $\epsilon$, independent Bandits, and interaction Bandits. We also find the proposed algorithm's performance is the most robust to changes in the across-page interdependence strength.
\end{abstract}





\maketitle
\section{Introduction}
\label {sec:intro}

For marketing, we sometimes need to recommend content for multiple pages in sequence. At each page, users can choose to continue the process or leave. Use cases like these require sequential decision making process since each page has a set of content candidates and the algorithm needs to make recommendation considering interdependence between content on successive pages. Different from general sequential decision making process, it is a simpler flow where customers per seeing recommended content on each page can only move forward in the process or drop from it. We define problems like this as sequential decision making in linear--flow. We observe this type of problems in customer engagement beyond content recommendation. For instance, it is common to take a series of nudges, such as promotion email and limited-time discount offer, to increase customer's activity. We can consider customers enter the workflow when their interactions drop below a threshold, and leave the flow with reward as $+1$ if their interactions increase to a healthy level or $-1$ if customers become inactive. 


In this paper, we formulate sequential decision making process in linear-flow as a Markov Decision Process (MDP). We model the transition probability matrix with contextual Bayesian Bandits \cite{MVT}, use Thompson Sampling (TS) as the exploration strategy, and apply exact Dynamic Programming (DP) to solve the MDP. Modeling transition probability matrix with contextual Bandits makes it convenient to specify actions' incompatibility, when certain action pairs at different steps are ineligible. It also makes it straightforward to interpret interdependence strength between actions in two successive situations. Our algorithm is a realization of the Posterior Sampling for Reinforcement Learning (PSRL) framework \cite{PSRL, osband2017posterior} with one difference. PSRL framework assumes each (state, action) pair is modeled separately while we model the pairs at a step using one Bayesian Bandit model. We present our algorithm in section \ref{methods}. We use simulator to evaluate the proposed algorithm comparing to Q-learning with $\epsilon$-greedy and decreasing $\epsilon$, independent Bandits, and interaction Bandits. The evaluation results are documented in section \ref{simulationResults}. We then conclude the paper in section \ref{conclusion}. In the remainder of this paper, we relate writing to content recommendation applications while the algorithm works for any linear-flow problems.  

\section{Methods}
\label {sec:methods}




\label{methods}
\subsection{Problem Formulation}
We model sequential decision making in linear flow as an MDP. Let $D$ be number of the pages that content recommendation is needed. For each page $i$, there are $N_{i}$ content candidates that we can choose from. The selected content can be denoted as $a_{i,n_i}$ where $n_i \in \{ 1, 2, \cdots, N_{i} \}$. For simplicity, we will denote the content shown on page $i$ as $a_{i}$ unless the notation $a_{i,n_i}$ is needed to distinguish between different content for page $i$. The flow starts with state $S_{1, X}$ where $X$ is customer feature vector. After action $a_{i}$ is selected to show to customer with features $X$ on page $i$, there are two possible states: $S_{i+1, X, a_{i}}$ and $S_{Exit}$. The only exception is for the last page, page $D$, the two possible states after action $a_{D}$ are: $S_{End}$ and $S_{Exit}$. $S_{i+1, X, a_{i}}$ means customers move forward in the process to the next page after seeing $a_{i}$. $S_{Exit}$ means customers drop from the process without reaching the end and $S_{End}$ means customers reached the end. We denote $R_{i}$ and $G_{i}$ as short--term and long--term rewards after action ${a_{i}}$ respectively. The goal is to optimize $G_1$. Customers are at $S_{1,X}$ after they trigger the flow. $X$ may include customer features, the upstream channel that customers land from, etc. Customers get to $S_{Exit}$ if they drop from the process at any page leading to $R_{i} = 1$. Customers are in $S_{i+1, X, a_{i}}$ if they move on in the process after marketing content $a_{i}$. The short-term reward is 0 from entering $S_{i+1, X, a_{i}}$. $S_{End}$ is another termination state other than $S_{Exit}$. Customers reach this state if they decide not to take the offer after seeing content on the last page. This leads to $R_{D} = 0$.  

\subsection{Transition Probability Matrix Modeling}
We use contextual Bandit to model and update transition probability matrix. As our rewards are binary and there can be only two outcomes per action, we select BLIP \cite{BLIP} as the Bandit algorithm. The transition probability is formulated through a Probit link function. After showing page $i$, the probability of customers moving forward in the process and dropping from the process is formulated as in Equations \ref{eq:1} and \ref{eq:2} respectively. By using this formulation, we model only interdependence between two successive pages. For page 1, $a_{i-1}$ is dropped from the subscripts but the rest of the equations remain the same.

\begin{equation}
P(S_{t+1} = S_{i+1, X, a_{i}}|S_{t} = S_{i, X, a_{i-1}}, a_{t} = a_{i}) = \Phi(-1 * \frac{B_{X, a_{i-1}, a_{i}}^TW_{i}}{\beta}) \label{eq:1}
\end{equation}

\begin{equation}
P(S_{t+1} = S_{Exit}|S_{t} = S_{i, X, a_{i-1}}, a_{t} = a_{i}) = \Phi(\frac{B_{X, a_{i-1}, a_{i}}^TW_{i}}{\beta}) \label{eq:2}
\end{equation}

In the formula, $\Phi$ is cumulative distribution function for standard normal distribution. $\beta$ is a scaling hyperparameter. $W_{i}$ are the weights quantifying feature contributions to the utility function. $B_{X, a_{i-1}, a_{i}}$ is the final vector that we will learn weights for and formed by combining $X$, $a_{i-1}$ and $a_{i}$ (usually with interaction terms). $B_{X, a_{i-1}, a_{i}}$ is the reason why we can easily model content incompatibility. To explain this, let's first assume we have one feature $x$ for $X$, $a_{i-1}$ has two candidates: $a_{i-1,1}$ and $a_{i-1,2}$, and $a_{i}$ has three candidates: $a_{i,1}$, $a_{i,2}$ and $a_{i,3}$. We assume $a_{i,3}$ cannot be shown if $a_{i-1,1}$ is shown to customers. We may set $B_{a_{i-1}, a_{i}}$ as in Equation \ref{eq:3} where we have $1$ for intercept and contextual feature interacting with action terms. We recognize content incompatibility between $a_{i-1,1}$ and $a_{i,3}$ by simply omitting interaction term $x \cdot a_{i-1,1} \cdot a_{i,3}$ in the equation. In addition, we can interpret interdependence strength between pages by checking the weights for the interaction terms across pages.


\begin{equation}
  \begin{split}
    B_{X, a_{i-1}, a_{i}}^T &= (1, x \cdot a_{i,1}, x \cdot a_{i,2}, x \cdot a_{i,3}, \\
    &a_{i-1,1} \cdot a_{i,1}, a_{i-1,1} \cdot a_{i,2}, \\
    &a_{i-1,2} \cdot a_{i,1}, a_{i-1,2} \cdot a_{i,2}, a_{i-1,2} \cdot a_{i,3}) \label{eq:3} \\
  \end{split}
\end{equation}

\begin{algorithm*}
\begin{algorithmic}[1]
\State Init $\mathscr{D}_1 = \mathscr{D}_2 = \cdots = \mathscr{D}_D = \oldemptyset$
\For{$k = 1,\cdots, K$}
    \State Receive context $x_k$ and Initiate $\E[G|a_D] = 0$
    \State Sample $W_{1}, \cdots, W_{D}$ from the posterior $P(W_1|\mathscr{D}_1),\cdots, P(W_D|\mathscr{D}_D)$ \algorithmiccomment{Thompson Sampling}
    \For{$i = D, D - 1, \cdots, 2$} \algorithmiccomment{Line 5 - 11: Exact Dynamic Programming}
        \For{ $a_{i - 1}$ in content pool of page $i - 1$}: 
                \State $a_i^*(a_{i-1}) = \underset{a_{i}}{\argmax} \big\{ \E[R|a_{i}, a_{i-1}, x_k; W_{i}] + (1 - \E[R|a_{i}, a_{i-1}, x_k; W_{i}]) * \E[G|a_{i}] \big\}$ \label{eq:eq2}
                \State $\E[G|a_{i - 1}] = \E[R|a_i^*(a_{i-1}), a_{i-1}, x_k; W_{i}] + (1 - \E[R|a_i^*(a_{i-1}), a_{i-1}, x_k; W_{i}]) * \E[G|a_i^*(a_{i - 1})]$ \label{eq:eq3}
        \EndFor
    \EndFor
\State $a_1^* = \underset{a_{1}}{\argmax} \big\{ \E[R|a_{1}, x_k; W_1] + (1-E[R|a_{1}, x_k; W_1])\E[G|a_{1}]\big\} $
\State Display layout $\{a_1^*,a_2^*(a_1^*),\cdots, a_{D}^*(a_{D-1}^*)\}$ and observe reward $R_{1,k}, \cdots, R_{D,k}$ \algorithmiccomment{Take actions and collect feedbacks}
\State Update $\mathscr{D}_1 = \mathscr{D}_1 \cup (x_{k}, a_{1k}, R_{1k})$ and update $\boldsymbol{\mu}_{W_1}$, $\boldsymbol{\nu}_{W_1}$ \algorithmiccomment{BLIP Updating. Formula (7, 8) in \cite{BLIP}}
\State Update $\mathscr{D}_i = \mathscr{D}_i \cup (x_{k}, a_{ik}, a_{i-1,k}, R_{i,k})$ and update $\boldsymbol{\mu}_{W_i}$, $\boldsymbol{\nu}_{W_i}$ for $i\in \{2,3,\cdots,D\}$ if $a_{ik}$ was presented
\EndFor
\end{algorithmic}
\caption{MDP with Bandits}
\label{alg:seqTS}
\end{algorithm*}

We assume the weights $W_{i}$ follow Normal distribution parameterized by a mean vector $\boldsymbol{\mu}_{W_i}$ and variance vector $\boldsymbol{\nu}_{W_i}$. We update the distribution from prior to posterior by enforcing Normal distribution as the distribution type and obtaining the distribution parameters by minimizing KL-divergence from the approximate normal distribution and the exact distribution.  

\subsection{RL Policy and Policy Updating}
We propose Algorithm \ref{alg:seqTS} \textit{MDP with Bandits} to solve the formulated MDP. $\mathscr{D}_d$ denotes the dataset for updating contextual Bandit for Page $d$. $k$ denotes the $k^{th}$ impression. $a_i^*(a_{i-1})$ denotes the optimal action for page $i$ given content on page $i-1$. The complexity of the algorithm is $O(DN^2K)$ assuming number of contents on each page is $N$ for discussion simplicity. For each observation, we sample weights from their posterior distribution using TS for calculating transition probability matrix. We then employ exact DP to allocate the optimal series of content to show on each page given the weights. Per reward signal availability, we update the weights distributions and the loop repeats. If we get the feedback signal in batches, the algorithm alters on Lines 11 and 12 where we would have the updates at batch level instead of observation level.

\section{Simulation}
\label {sec:results}
\label{simulationResults}
Using simulation, we compare the proposed \textit{MDP with Bandits} algorithm to three other algorithms: independent Bandits, interaction Bandits and Q-learning. Independent Bandits algorithm assumes no interdependence across pages. TS for each Bandit model is conducted separately for each page. Interaction Bandits algorithm models interdependence in two successive pages by using the content shown in a previous page to construct features in the Bandit formulation. Other than the feature construction, TS for each Bandit model in the interaction Bandits algorithm is also conducted separately. As shown in Table \ref{table:alg_set}, in \textit{MDP with Bandits}, we use short-term reward $R_i$ as the feedback signal for Bandits while independent and interaction Bandits use the long-term reward $G_i$ as the feedback signal. For notation simplicity, we use the tide sign $\sim$ to introduce the linear model form used for Equations \ref{eq:1} and \ref{eq:2} with Probit link function assumed as default. The Bandits forms only involve content features to highlight the action interaction. \textit{MDP with Bandits} and interaction Bandits have the same linear model form. Lift in performance from \textit{MDP with Bandits} (if any) directly quantifies the benefit of modeling the process as an MDP instead of a series of Bandits. For Q-learning, we follow the vanilla setup: we model the content shown in the previous page as the state and update the Q-Table based on Bellman Equation. We set the learning rate to be 0.05 and discount factor to be 1. The policy is based on $\epsilon$-greedy with $\epsilon$ linearly decreases from 0.05 (2nd batch) to 0.01 (last batch).

\begin{table*}[h]
\centering
\caption{Algorithms in Comparison}
\label{table:alg_set}
\begin{tabular}{|l|l|l|}
\hline
Algorithm & Linear model form & \# parameters \\
\hline
MDP with Bandits &  Page 1: $R_1 \sim a_1$; Page $i$: $R_i \sim a_i + a_{i - 1} + a_{i - 1}:a_i$ for $i\in 2,\cdots,D$  & $O(DN^2)$ \\ 
\hline
Interaction Bandits & Page 1: $G_1 \sim a_1$; Page $i$: $G_i \sim a_i + a_{i - 1} + a_{i - 1}:a_i$ & $O(DN^2)$ \\ 
\hline
Independent Bandit & Page 1: $G_1 \sim a_1$; Page $i$: $G_i \sim a_i$ & $O(DN)$\\ 
\hline
\end{tabular}
\end{table*}

\subsection{Simulated Data}
We generate simulation data assuming Probit link function between short-term reward $R_{i}$ and $B_{X, a_{i-1}, a_{i}}^TW_{i}$ based on equations \ref{eq:1} and \ref{eq:2}, with $\beta$ set as $1 + \alpha_1 + \alpha_c + \alpha_2$ and the linear model forms for $B_{X, a_{i-1}, a_{i}}^TW_{i}$ specified as in equation \ref{eq:sim_first} for first page and equation \ref{eq:sim} for the other pages. Similar to \cite{MVT}, we use the $\alpha_1$, $\alpha_c$ and $\alpha_2$ to separately control the importance of content to recommend on the current page, content already shown on the previous page, and the interaction between them. For each simulation run, the ground truth weights $W_i$ are sampled independently from a normal distribution with mean $0$ and variance $1$ except for $w_i^0$. Mean for $w_i^0$ is set as $\Phi^{-1}(x)\beta$ where $x$ reflects the average success rate. With a larger value for $\alpha_2$, interdependence across pages is set stronger.

\begin{equation}
B_{X, a_{1}}^TW_{1} = w^0_1 + \alpha_1 \sum_{n_1=1}^{N_1} w_{n_1}^1 a_{1,n_1} + \alpha_1 \sum_{x} w_{x}^1 x \label{eq:sim_first}
\end{equation}

\begin{equation}
	\begin{split}
		B_{X, a_{i-1}, a_{i}}^TW_{i} &= w^0_i + \alpha_1 \sum_{n_i=1}^{N_i} w_{n_i}^1 a_{i,n_i} + \alpha_1 \sum_{x} w_{x}^1 x + \\
    &\alpha_c \sum_{n_{i-1}=1}^{N_{i-1}} w_{n_{i-1}}^c a_{i-1,n_{i-1}} + \\
    &\alpha_2 \sum_{n_i=1}^{N_i} \sum_{n_{i-1}=1}^{N_{i-1}} w_{n_i,n_{i-1}}^2 a_{i,n_i} a_{i-1,n_{i-1}} + \\
    &\alpha_2 \sum_{n_i=1}^{N_i} \sum_{x} w_{n_i,x}^2 a_{i,n_i} x \quad \textrm{for} \quad  i\in 2,\cdots,D \label{eq:sim} \\
	\end{split}
\end{equation}

For each simulation, we sample the actual weights in equations \ref{eq:sim_first} and \ref{eq:sim} once and we calculate the long-term reward of each combination $(a_1, \cdots, a_D)$ as:

\begin{equation}
  \begin{split}
\E[G|W,&a_1,\cdots,a_D] = \\
    &\E[R_1 + (1 - R_1)R_2 + \cdots + \Pi_{i=1}^{D-1}(1 - R_i)R_D] \label{eq:expected_reward}
  \end{split}
\end{equation} 
where $R_i = \Phi(\frac{B_{X,a_{i-1}, a_{i}}^TW_{i}}{\beta}) \quad \textrm{for} \quad  i\in 2,\cdots,D$ and $R_1 = \Phi(\frac{B_{X, a_{1}}^TW_{1}}{\beta})$. 

The series of actions for each algorithm are selected by itself. Given sampled weights and selected actions, empirical rewards ($R$) are sampled from the data generation model. Actual reward $G_k$ is 0 if any $R_k$ is 0. Each simulation was run for $K = 14k$ time steps with updating batch size $1k$ to mimic a 2-week experiment. We ran $100$ simulations and report the averaged performance of each algorithm using batch size normalized cumulative regret:
\begin{equation}
\frac{1}{\# runs}\sum_{j=1}^{\# runs} \frac{1}{batchSize}\sum_{k = 1}^{K} (\E[G|W,a_{1,j}^*, \cdots, a_{D,j}^*] - G_{j,k}) \label{eq:regret}
\end{equation} 

Where $\E[G|W,a_{1,j}^*, \cdots, a_{D,j}^*]$ is the best expected long-term reward for $j^{th}$ run and $G_{j,k}$ is the long-term reward for $j^{th}$ run and $k^{th}$ time step.

\subsection{Simulation Results}

\begin{figure}[tb]
\centering
\begin{minipage}[b]{.48\textwidth}
  \centering
  \includegraphics[width=1.0\textwidth]{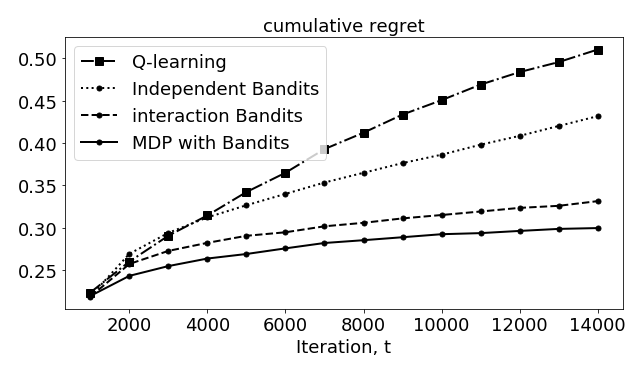}
  \caption{Example run of algorithms on simulated data with $\alpha_1=1, \alpha_c=1, \alpha_2=2$~~~~~~~~~~~~~~~~~~~~~~~~~~~~~~~~~~~~~~~~~~~~~~}
  \label{fig:darwinStrength2}
\end{minipage}%
\hfill
\begin{minipage}[b]{.48\textwidth}
  \centering
  \includegraphics[width=1.0\textwidth]{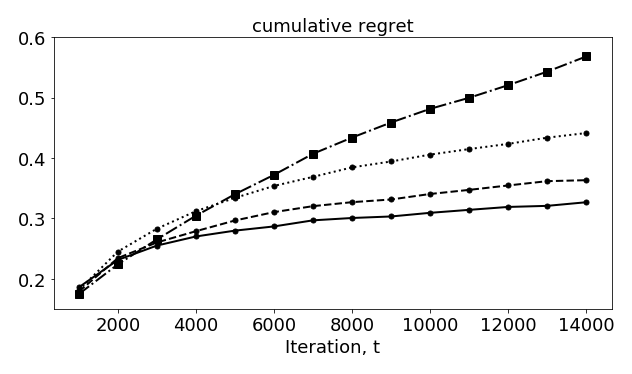}
  \caption{Example run of algorithms on simulated data with $\alpha_1=1, \alpha_c=1, \alpha_2=2$ and 1 categorical feature with 3 categories}
  \label{fig:darwinContext}
\end{minipage}
\end{figure}

\begin{figure}[tb]
\centering
\begin{minipage}[b]{.48\textwidth}
  \centering
  \includegraphics[width=1.0\textwidth]{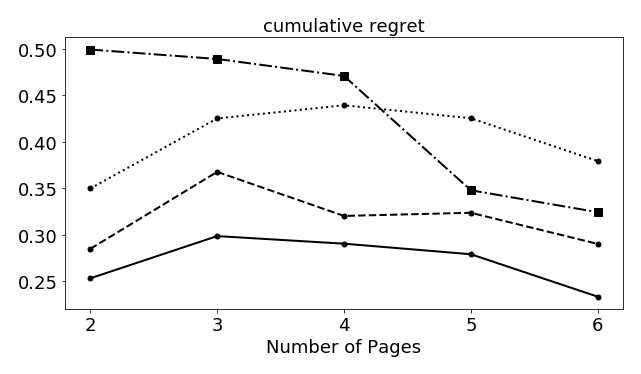}
  \caption{Algorithm performance as number of pages vary. $\alpha_1 = 1, \alpha_c = 1, \alpha_2 = 2$}
  \label{fig:darwinPage}
\end{minipage}%
\hfill
\begin{minipage}[b]{.48\textwidth}
  \centering
    \includegraphics[width=1.0\textwidth]{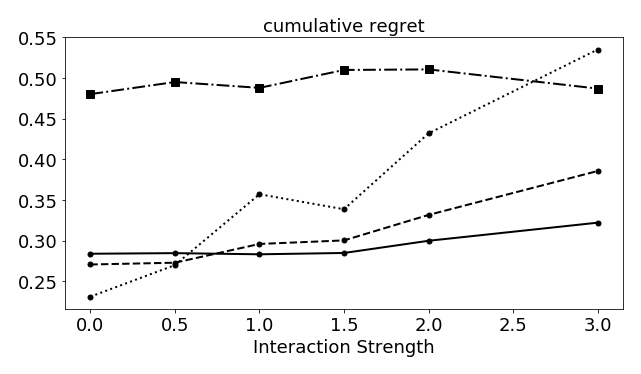}
  \caption{Algorithm performance as $\alpha_2$ is varied. $\alpha_1=1, \alpha_c=1$}
  \label{fig:darwinStrength}
\end{minipage}
\end{figure}

We first compare the performance with $\alpha_1=1, \alpha_c=1, \alpha_2=2$ in both non-contextual and contextual use case. In the contextual case, we include $1$ categorical feature with $3$ equally-likely categories. We assume $3$ pages and each page has $3$ content candidates. Results are shown in figures \ref{fig:darwinStrength2} and \ref{fig:darwinContext}. The proposed algorithm converges faster than all the other three. Also, it has the lowest cumulative regret followed by interaction Bandits. This means when there is considerable interdependence across pages, modeling interactions in the transition probability reduces regrets and modeling the process as an MDP instead of separate Bandits further lifts the performance. On the other hand, bandit algorithms with TS generally outperform Q-learning with $\epsilon$-greedy, given its efficiency in exploration.

We then take a look at the impact of number of pages on performance, shown in figure \ref{fig:darwinPage}. The number of pages varies from $2$ to $6$ and that of content combinations $N^D$ varies from $9$ to $729$. \textit{MDP with Bandits} consistently outperforms the other three algorithms. The performance advantage over Q-learning shrinks with increasing page number.

To test the model's sensitivity to interdependence strength, we conduct a set of simulation runs with $\alpha_2$ increasing from 0 to 3 while fixing $\alpha_1$ and $\alpha_c$ at $1$. Results are shown in figure \ref{fig:darwinStrength}. When $\alpha_2$ is $0$, independent bandits algorithm outperforms the other two bandit algorithms as both interaction Bandits and \textit{MDP with Bandits} require traffic to learn the weights for interaction terms should actually be set as $0$. As $\alpha_2$ increases, \textit{MDP with Bandits} starts to outperform. As Q-learning directly learns the Q function by Bellman equation, interaction strength has minor influence to its performance. Overall, the performance for \textit{MDP with Bandits} is consistently better than all other algorithms when interaction strength is greater than 1.

When scanning through figures \ref{fig:darwinStrength2}, \ref{fig:darwinContext}, \ref{fig:darwinPage}, \ref{fig:darwinStrength}, we find interaction Bandits is the 2nd best when \textit{MDP with Bandits} is the best and its performance is comparable to that of \textit{MDP with Bandits}. When interaction strength is small, it can perform better than \textit{MDP with Bandits}, as at the left bottom of figure \ref{fig:darwinStrength}. In addition, similar to \textit{MDP with Bandits}, the cumulative regrets of interaction Bandits always level out as in figures \ref{fig:darwinStrength2}, \ref{fig:darwinContext}. These mean interactive Bandit, where separate contextual Bandits with long-term reward and action shown on the previous page as feature are built to model each page, is a good surrogate modeling strategy to a multi-step RL and always converges to the optimal actions in the case of stationary reward. This is because that interactive Bandit samples the best action for each page under the expected rewards of all the trajectories after the action. With improving performance estimation regarding each trajectory along training, the algorithm is able to learn the trajectory with the highest long-term reward. We do not though consistently observe convergence for independent Bandits and Q-learning. We observe independent Bandits got stuck in local optimal actions that maximize short-term but not long-term reward for some simulation runs, even if it is using the long-term reward. For Q-learning, theoretically it is guaranteed to converge, but with the conditions that the learning rate needs to be diverged and each state-action pair must be visited infinitely often \cite{watkins1992q}. In practice, it’s not trivial to tune the parameters and the convergence may not be obtained.

\section{Conclusions}
\label {sec:conlcusion}
\label{conclusion}
We proposed a RL algorithm with posterior sampling for recommendation in multi-step linear--flow. We set the problem up as an \textit{MDP with Bandits} to model transition probability matrix. At policy inference time, we use TS to sample the transition probabilities and allocate the best series of actions with exact dynamic programming. The algorithm is able to leverage TS's efficiency in balancing exploration and exploitation and Bandit's convenience in modeling actions' incompatibility. In the simulation study, we observe the proposed algorithm outperforms Q-learning with $\epsilon$-greedy, independent Bandits, and interaction Bandits, under different scenarios as long as there is reasonable interdependence strength across pages. We also see the proposed algorithm is the most robust to changes in the interdependence strength. We often face dynamic environments where actions and interdependence between sequential actions are subject to change. Robustness to changes is thus a good to have. On the other hand, we find interactive Bandit, where separate contextual Bandits with long-term reward and action shown on the previous page as feature are built to model each page, is a good surrogate modeling strategy to a multi-step. Just like the proposed algorithm, it always converges to the optimal actions in the case of stationary reward. This is evidence that when the long-term impact of actions is significant, Bandits if set up properly may not be the best formulation, but can still be a good baseline.   

Our work is a start in using Bandit (1-step RL) to solve stateful RL (multi-step RL). More efforts are in need to generalize the application. For instance, one desired algorithm extension is to use Bandit to model a multi-stage flow where customers can move forward with more than two options after seeing an action. In terms of algorithm evaluation, we seek to leverage logged data for counterfactual evaluation in addition to simulation analysis \cite{dudik2011doubly, strehl2010learning}.  

\section*{Acknowledgments}
\label {sec:acknowledgments}
We would like to thank Lihong Li and Shipra Agrawal for their valuable comments and helpful discussions.

\bibliography{darwin}
\bibliographystyle{icml2020}

\end{document}